\documentclass[runningheads]{llncs}

\usepackage[misc]{ifsym}
\usepackage{graphicx, csquotes}
\graphicspath{{figures/}}
\usepackage{amsmath,amssymb,mathtools} 
\usepackage{tikz, ifthen}
\usetikzlibrary{calc, intersections, fit, positioning, matrix}

\usepackage{hyperref}

\newtheorem{defi}{Definition}
\newcommand{\KL}{\text{KL}}
\newcommand{\tr}{\operatorname{tr}}

\newcommand{\N}{\mathcal{N}}
\newcommand{\W}{\mathcal{W}}

\newcommand{\BigO}{\mathcal{O}}

\makeatletter
\newcommand\thefontsize{\f@size pt}
\makeatother

\newcommand{\abc}[1]{(\textsf{\bfseries #1})}

\definecolor{imprsdb}{RGB}{035, 127, 154} % IMPRS dark blue
\definecolor{TUred}{RGB}{141,45,57}
\definecolor{TUdark}{RGB}{55,65,74}
\definecolor{TUgold}{RGB}{174,159,109}
\definecolor{TUgray}{RGB}{175,179,183}

\newboolean{anonymize}
\setboolean{anonymize}{False}

\begin{document}

\title{Wasserstein \textit{t}-SNE}
\toctitle{Wasserstein t-SNE}
\tocauthor{Fynn~Bachmann}

\ifthenelse{\boolean{anonymize}}
    {%TRUE -> anonymize authors
        \author{Anonymous Authors}
        \authorrunning{Anonymous Authors}
        \institute{Anonymous Institute}
        \def\pypipackage{at [PyPi link]}
        \def\githubpaper{at [GitHub link]}
        \def\githubpackage{at [GitHub link to be added after review]}
    }{%FALSE -> show authors
        \author{	Fynn Bachmann\inst{1,2}{\Letter}%\orcidID{0000-0001-9571-5671} 
        	\and    Philipp Hennig\inst{2}%\orcidID{0000-0001-7293-6092}
        	\and	Dmitry Kobak\inst{2}%\orcidID{0000-0002-5639-7209}
        	}
        \authorrunning{F. Bachmann et al.}
        \institute{University of Hamburg, Germany\\
        		\email{fsvbach@gmail.com}
         	\and 	University of Tübingen, Germany\\
         		\email{\{philipp.hennig, dmitry.kobak\}@uni-tuebingen.de}}
        \def\pypipackage{on PyPi~\href{https://pypi.org/project/WassersteinTSNE/}{\textsf{WasserteinTSNE}}}
        \def\githubpackage{on GitHub at~\href{https://github.com/fsvbach/WassersteinTSNE}{\textsf{fsvbach/WassersteinTSNE}}}
        \def\githubpaper{on GitHub at~\href{https://github.com/fsvbach/wassersteinTSNE-paper}{\textsf{fsvbach/wassersteinTSNE-paper}}}
    }
    
\maketitle             
%
%%% Abstract
\begin{abstract}
Scientific datasets often have hierarchical structure: for example, in surveys, individual participants (samples) might be grouped at a higher level (units) such as their geographical region. In these settings, the interest is often in exploring the structure on the unit level rather than on the sample level. Units can be compared based on the distance between their means, however this ignores the within-unit distribution of samples. Here we develop an approach for exploratory analysis of hierarchical datasets using the Wasserstein distance metric that takes into account the shapes of within-unit distributions. We use $t$-SNE to construct 2D embeddings of the units, based on the matrix of pairwise Wasserstein distances between them. The distance matrix can be efficiently computed by approximating each unit with a Gaussian distribution, but we also provide a scalable method to compute exact Wasserstein distances. We use synthetic data to demonstrate the effectiveness of our \textit{Wasserstein \mbox{$t$-SNE}}, and apply it to data from the 2017 German parliamentary election, considering polling stations as samples and voting districts as units. The resulting embedding  uncovers meaningful structure in the data. 
% We argue that Wasserstein $t$-SNE can be a useful tool for visualizing hierarchical data in other domains as well.

\end{abstract}

%%% This is the main part of the paper
\section{Introduction}
\label{sec:INTRO}

%%% FIELD
%Dimension reduction algorithms are established to visualize structure or to reduce the complexity of datasets~\cite{burges2010dimension}. Usually the datapoints are high dimensional feature vectors, for example images~\cite{chan2019gpu} or cells~\cite{kobak2019art}, that are  embedded into the 2D-plane while keeping the original structure. This can reveal meaningful clusters within the dataset to motivate further analysis or serve as an overview for exploration.

\begin{figure}[t]
	\input{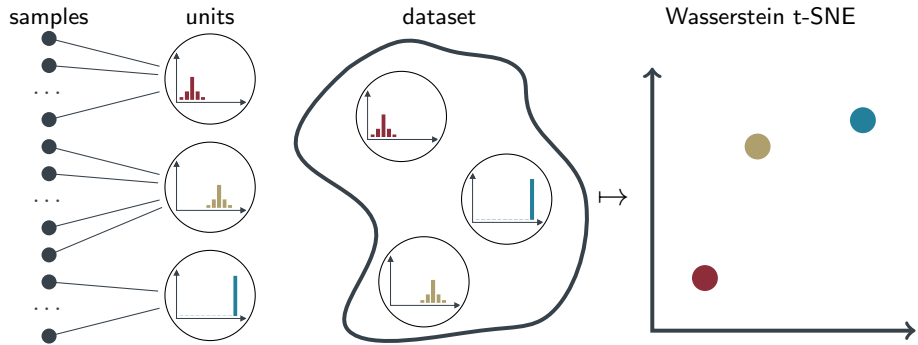}	
	\caption{Hierarchical data. Individual \textit{samples} can be grouped into \textit{units}. Each unit forms a probability distribution over its samples.  In our \textit{Wasserstein $t$-SNE} approach, units in the dataset are compared using the Wasserstein metric to construct a pairwise distance matrix, which is then  embedded in two dimensions using the $t$-SNE algorithm. Units with similar probability distributions end up close together in the 2D embedding.}
	\label{fig:WTcomic}
\end{figure}

%%% DATA
We consider dimensionality reduction for the purpose of data visualization, for the situation in which each `data point' is a probability distribution, or a set of samples from it. This situation naturally arises when the data have \emph{hierarchical structure}, i.e. the individual samples can be grouped at a higher level. Throughout this work we will use the word `unit' to refer to this grouping level; for each `unit' there is a number of `samples' in the data (Figure~\ref{fig:WTcomic}). For example, in a social science survey, participants can be seen as samples and their countries of origin can be seen as units. For exploratory analysis, the interest may often be in the relationships between units (countries), rather than samples (participants). %Instead of embedding the individual participants we are interested in visualizing the structure of the regions. Considering them as probability distributions their comparison should not only include means, but also covariances and  possibly higher moments as well. This is especially relevant if the shape of a distribution depends on the region, i.e., when certain features correlate in some units but anti-correlate in others. 

%%% FIGURE to TOP of 2nd PAGE

%%% DEFINE PROBLEM
%The goal of any dimension reduction algorithm is to preserve as much information about the distances in the high dimensional space as possible. 
A common approach for data exploration is to visualize the dataset as a 2D embedding,  using dimensionality reduction algorithms such as PCA, MDS, {$t$-SNE}~\cite{van2008visualizing} or UMAP~\cite{mcinnes2018umap}. These algorithms are designed to get vectors as input, and compute pairwise distances (e.g. Euclidean) between the input vectors. However, when analyzing units in a hierarchical dataset, each single unit forms an entire probability distribution over its samples, and cannot be represented by one vector.  A simple approach would be to collapse all within-unit distributions to their means, and then apply any standard dimensionality reduction algorithm. However, this procedure can loose important information, particularly when some of the units share the same mean but have different shape. 

%%% PROPOSE SOLUTION
Here we propose to use the Wasserstein metric~\cite{kantorovich1939mathematical} to compute pairwise distances between units. The Wasserstein distance has got recent attention in applications to Generative Adversarial Networks~\cite{arjovsky2017wasserstein} or discriminant analysis~\cite{flamary2018wasserstein} where it was used to compare probability densities with different support. The Wasserstein metric is convenient because there exists a closed-form solution for Gaussian distributions~\cite{dowson1982frechet}. Using the Gaussian approximation, it is possible to efficiently construct the pairwise distance matrix between units in a hierarchical dataset. This distance matrix can then be used for downstream analysis, such as clustering or dimensionality reduction. Our focus here will be on $t$-SNE embeddings.

%%% SUMMARY
In the first part of this work we use simulated data to demonstrate the effectiveness of our \textit{Wasserstein $t$-SNE}. In the second part we apply the same method to real-world data, in particular the data from the 2017 German parliamentary election. Here, samples correspond to polling stations while the units correspond to voting districts. We use the Gaussian approximation but also compute the exact Wasserstein distances, using an efficient linear programming approach. 
%%Our objective is not to compare different dimensionality reduction algorithms, but to show that using Wasserstein metric can be useful for embedding hierarchical data. 

The Python implementation of Wasserstein $t$-SNE is available \githubpackage{} and as a package \pypipackage{}. The analysis code reproducing all figures in this paper can be found \githubpaper{} together with the analyzed data.
\section{Methods}

\subsection{$t$-SNE}
\label{sec:TSNE}

{T-distributed stochastic neighbor embedding} ($t$-SNE)~\cite{van2008visualizing} is a dimensionality reduction algorithm used in many scientific fields to find structure in datasets. %While it was originally proposed to embed vectors, it works as well with a distance matrix as input. In this section we describe the latter approach since the units we embed later are probability distributions which cannot be represented in Euclidean space. 
The main idea of $t$-SNE is to arrange points in a low-dimensional space, such that their pairwise similarities (affinities) are similar to those in the high-dimensional space. In particular, if $P$ and $Q$ are the affinity matrices of the data and embedding respectively, $t$-SNE minimizes their Kullback-Leibler divergence 
\begin{equation*}
	\KL(P\|Q):=\sum_{i j} P_{i j} \log \frac{P_{i j}}{Q_{i j}}.
\end{equation*}

The affinity matrix $P$ is constructed from the pairwise distances $d_{ij}$ by Gaussian kernels with bandwidth $\sigma_i$ 
\begin{equation*}
	P_{j\mid i} = \frac{\exp( - d_{ij}^2 / 2\sigma_i^2)}{\sum_{k \neq i} \exp(-d_{ik}^2 / 2\sigma_i^2)}
\end{equation*}
such that all perplexities of the conditional distributions equal some predefined value. In most $t$-SNE implementations this parameter defaults to 30 which we leave untouched in our experiments. As a reminder, if $p(x)$ is a discrete probability density function, the perplexity of $p$ is given by 
\begin{equation*}
	 \mathcal P(p) := 2^{H(p)}=\prod_x p(x)^{-p(x)}.
\end{equation*}
The affinity matrix $P$ is then symmetrized by
\begin{equation*}
 	P_{ij} = \frac{P_{j\mid i} + P_{i\mid j}}{2n}.
\end{equation*}
In the low-dimensional space, the affinity matrix $Q$ is based on the pairwise distances between the embedding vectors $\mathbf{y}_{i}$, using the $t$-distribution kernel:
\begin{equation*}
	Q_{i j}=\frac{\left(1+\left\|\mathbf{y}_{i}-\mathbf{y}_{j}\right\|^{2}\right)^{-1}}{\sum_{k \neq l}\left(1+\left\|\mathbf{y}_{k}-\mathbf{y}_{l}\right\|^{2}\right)^{-1}}.
\end{equation*}

The $t$-SNE algorithm minimizes the Kullback-Leibler divergence $\KL(P\|Q)$ with respect to the coordinates $\mathbf{y}_{i}$. The embedding is initialized randomly, or using another algorithm such as PCA~\cite{kobak2021initialization}. 
The optimization is done with gradient descent, i.e. the points move along the gradient until convergence. This results in a local minimum where no point can be moved without yielding a worse embedding. %However, this embedding is not unique. 

When interpreting $t$-SNE embeddings it is important to keep in mind that the algorithm puts emphasis on close points, i.e., similar data points are embedded close to each other. The opposite does not hold: points that are embedded far from each other are not necessarily far from each other in the original space. 
 
%The $t$-SNE algorithm comes with certain hyperparameters and is dependent on initialization. Its objective is non-convex, thus it has multiple local minima in which the algorithm can terminate. A $t$-SNE embedding is therefore not unique, i.e., it can have certain artifacts that occur with some initialization and not with others. 

%The main parameter that needs to be set is the perplexity $\sigma$. It defines the scale on which points are considered to be close. If the perplexity is very low, the structure will be very fine with lots of clusters since only few points are considered to be close. If the perplexity is high, all points might end up in the same cluster.

In this work we use the implementation of \textsf{openTSNE}~\cite{Policar731877}, and keep all parameters at their default values.

\subsection{Wasserstein metric}
\label{sec:WT}

The Wasserstein metric~\cite{kantorovich1939mathematical} is a natural choice to compare probability distributions. %A particular benefit is the sensitivity to the metric space on which their probability densities are defined. 
It can be used to compare densities which do not have the same support, as long as a distance measure of their support is given. The downside of the Wasserstein distance is its computational complexity, which is linked to optimal transport~\cite{villani2009optimal,memoli2011gromov}. 

\begin{defi}
	\label{def:WT}
		Let (M,d) be a metric space. The $p$-Wasserstein distance of two distributions $\mu$ and $\nu$ is  defined as
		$$W_p(\mu,\nu) := \left(\inf_{\gamma \in \Gamma(\mu,\nu)} \int_{M \times M} d(x,y)^{p} \mathrm{d} \gamma (x,y)\right)^{\frac {1}{p}}$$
		where $\Gamma$ is the set of all couplings of $\mu$ and $\nu$. 
	\end{defi}

In computer science the 1-Wasserstein metric is also known as \emph{Earth Mover's Distance}, because if one imagines the probability distributions as piles of earth, then $W_p(\mu,\nu)$ represents the minimal amount of work necessary to transfer this mass from $\mu$ to $\nu$. This intuition also explains why the probability distributions must be defined on a metric space $M$, because we have to measure how far two points are away from each other, i.e. how far the mass has to be transported.  

In general, the $p$-Wasserstein distance for continuous distributions is  hard to compute~\cite{villani2009optimal}. But there exists a closed-form solution for the $2$-Wasserstein metric for multivariate Gaussian distributions~\cite{dowson1982frechet} (also known as Fr\'echet Inception Distance~\cite{jung2021internalized}). If $\mu, \nu$ are two Gaussian distributions $\N_i(m_i, C_i)$ with means $m_i$ and covariance matrices $C_i$, the 2-Wasserstein distance between them is given by
\begin{align*}
	W_2(\mu, \nu)^{2} &= \left\|m_{1}-m_{2}\right\|_{2}^{2}+ \tr\left(C_{1}+C_{2}-2\left(C_{2}^{1 / 2} C_{1} C_{2}^{1 / 2}\right)^{1 / 2}\right)\\
	 &= \left\|m_{1}-m_{2}\right\|_{2}^{2}+ \tr\left(C_{1}+C_{2}-2\left(C_{2} C_{1} \right)^{1 / 2}\right).
\end{align*}
The first term here is the Euclidean distance between the means, while the second term defines a metric on the space of covariance matrices~\cite{dowson1982frechet}. By introducing a hyperparameter~$\lambda\in[0,1]$ we can put emphasis either on the means or on the covariances. We therefore propose a convex generalization of the 2-Wasserstein distance for Gaussians: %can therefore be written as
\begin{equation}\tag{$\star$}
	\label{eq:lambda}
	\tilde{W}(\mu, \nu)^{2}  = (1-\lambda) \cdot \left\|m_{1}-m_{2}\right\|_{2}^{2}+ \lambda\cdot  \tr\left(C_{1}+C_{2}-2\left(C_{2} C_{1} \right)^{1 / 2}\right).
\end{equation}
This reduces to the Euclidean distance between the means for $\lambda = 0$ and to the distance between covariance matrices for $\lambda = 1$. The 2-Wasserstein distance corresponds to $\lambda = 0.5$ (up to a scaling factor).
%As the $t$-SNE algorithm is indifferent to scaling the distances linearly, the additional factors won't change the embeddings. Throughout this work we use the $\lambda$-notation to reconcile all three cases into one method. 

% \subsubsection{Computational Complexity}

In closed form, the 2-Wasserstein distance between two Gaussians can be computed in polynomial time. The matrix multiplication and the eigenvalue decomposition (for taking the square root) have $\BigO(d^3)$ complexity, where $d$ is the number of features. If there are $n$ units in the dataset, the $n\times n$ pairwise distance matrix can be computed in $\BigO(n^2d^3)$ time. 

\subsection{Linear programming}
\label{sec:linprog}

\begin{figure}[t]
    \includegraphics{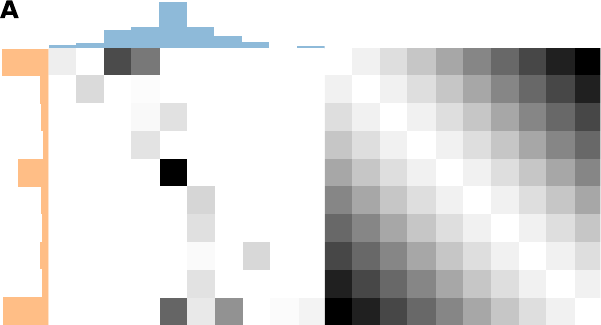}
    \includegraphics{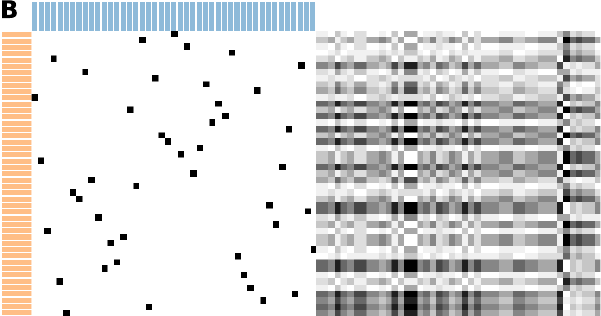}
	\caption{Wasserstein distance as a linear program. \abc{A} The optimal transport map $\gamma$ of two probability distributions $\nu$ (orange) and $\mu$ (blue) is shown. The heatmap represents the cost matrix $C$. \abc{B} The same distributions can be visualized as a collection of samples, which have different support. The distance between samples $\nu_i$ and $\mu_j$ is given in the cost matrix entry $C_{ij}$. The size of the optimization variable $\gamma$ is then upper bounded by the product of the sample sizes.}
	\label{fig:WTprogram}
\end{figure}

We are also interested in computing  exact Wasserstein distances without relying on the Gaussian approximation. Since real-life datasets contain discrete samples, this is possible by discretizing Definition~\ref{def:WT} to
$$W_p(\mu,\nu)^p := \min_{\gamma \in \Gamma(\mu,\nu)} \sum_{M \times M} d(m_i,m_j)^{p} \cdot \gamma(m_i,m_j),$$
which is equivalent to the following linear program~\cite{memoli2011gromov}:
\begin{equation}\tag{$\star\star$}
	\label{eq:linprog}
	\begin{array}{c|c}
		\mathbf{primal \ form:} & \mathbf{dual \ form:}\\
		\arraycolsep=.1cm
		\begin{array}{rrcl}
			\mathrm{minimize} \ & z & = & \ \mathbf{c}^T \mathbf{x}, \\
			\mathrm{so \ that} \ & \mathbf{A} \mathbf{x} & = & \ \mathbf{b} \\
			\mathrm{and}\  & \mathbf{x} & \geq &\ \mathbf{0}
		\end{array} &
		\arraycolsep=1mm
		\begin{array}{rrcl}
			\mathrm{maximize} \ & \tilde{z} & = & \ \mathbf{b}^T \mathbf{y}, \\
			\mathrm{so \ that} \ & \mathbf{A}^T \mathbf{y} & \leq & \ \mathbf{c}. \\ \\
		\end{array}
	\end{array}
\end{equation}
The vectorized matrix $\mathbf{c}$ defines the transport cost, i.e., $c_{ij}=d(m_i,m_j)^{p}$ represents the $L_p$-distance of the points $m_i$ and $m_j$, where $M$ is the discrete metric space on which the probability distributions are defined. The optimization variable $\mathbf{x}$ represents the vectorized transport plan $\gamma$ as in Figure~\ref{fig:WTprogram}A. Each entry of $\mathbf{x}$ must be non-negative. The constraint $\mathbf A\mathbf x=\mathbf b$ is set up such that it is satisfied if the marginals of $\gamma$ equal the densities $\mu, \nu$. The primal form in~(\ref{eq:linprog}) yields an explicit transport plan while the dual form has less variables and is faster. Due to the strong duality of a linear program the resulting solution $z$ is the same. In practice we therefore use the dual form to compute exact Wasserstein distances. 
%
%
%\subsubsection{Scalability}

Simplex algorithms and interior-point methods can solve real-world linear programs with a unique solution in polynomial time~\cite{huangfu2018parallelizing}. However, the exact complexity depends on the constraint matrix in the problem formulation. In general, the runtime of a linear program depends on the size of the optimization variable. In our case this is given by the product of the support sizes of the two discrete probability distributions. Figure~\ref{fig:WTprogram}A provides an example of two one-dimensional distributions, defined on the same support of size 10 (which could e.g. be a ten-point rating scale from 1 to 10). Both probability mass functions have $10$ values so the resulting optimization variable  $\gamma$ has $10{\times}10 = 100$ entries. While this linear program is easily solvable, the problem becomes computationally hard if we add additional feature dimensions (for example, if we add another ten-point feature, each probability density will become a two-dimensional mass function over 100 values, so then $\gamma$ has length 10,000). The number of variables in the transport map therefore grows exponentially with the number of features, thus this approach is intractable for datasets with many features.

Instead, we reduce the probability densities to the subspace where samples have actually been observed, rather than comparing distributions on the complete space $M$. That is, we consider both distributions uniformly distributed over their samples (Figure~\ref{fig:WTprogram}B). The marginal distributions $\mu$ and $\nu$ in Figure~\ref{fig:WTprogram}B therefore become uniform distributions with supports of size $n$ and $m$ respectively, where $n$ and $m$ are the two sample sizes. The size of the optimization variable $\gamma$ now becomes upper bounded by $nm$ regardless of the number of features. The cost matrix is given by the pairwise $L_p$-distance between all samples, which can be computed efficiently. We are not aware of a prior use of this shortcut, which however is only applicable when the number of samples is not large.

\begin{figure}[t]
	\centering
	\includegraphics{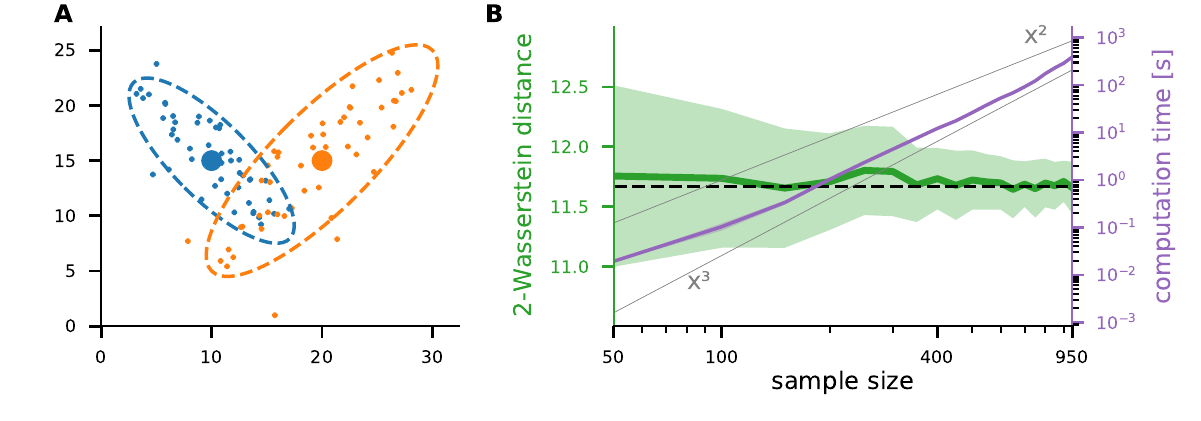}
	\caption{Computation time and accuracy. \abc{A} Two multivariate Gaussian distributions with $50$ samples each. \abc{B} The Wasserstein distance between the two probability distributions is computed using a different number of samples. The ground-truth distance is obtained by the closed-form solution and is shown with the dashed black line. The Wasserstein distance estimates using our linear program approach are shown in green (mean and standard deviation over $50$ repetitions). The purple line shows the average runtime.}
	\label{fig:WTcomputation}
\end{figure}

A way to see that both approaches are equivalent is to consider the constraints $\mathbf x \geq 0$ and $\mathbf A\mathbf x = \mathbf b$. When the number of features is large while the sample size is small, the sample density at most support values  will have zero probability mass, because no sample has been observed at that point. Since the rows and the columns in the transport plan must sum to the marginal distributions, each entry with zero probability mass forces the corresponding row or column to be empty. Therefore these entries can be left out in the problem formulation and the size of the optimization variable is upper bounded by $nm$. 
%In Figure~\ref{fig:WTprogram}B we show this approach. The probability distributions are the same as in Figure~\ref{fig:WTprogram}A, but we now consider them to be uniform over their samples.  The advantage of this formulation is that it is now indifferent to the number of features, since we can compute the pairwise Euclidean distance of the samples regardless their dimension. 

One consequence of this approach is that the samples no longer need to come from a discrete distribution, and indeed we can use the same approach to compute the exact Wasserstein distance between the samples coming from two Gaussian distributions (Figure~\ref{fig:WTcomputation}). To demonstrate that, we chose a pair of two-dimensional Gaussian distributions with Wasserstein distance $d_W=11.7$, where the Euclidean distance between the means is $d_E=10.0$ and the distance between the covariances is $d_C=6.0$ (Figure~\ref{fig:WTcomputation}A). As the sample size grows from 50 to 1000, the solution of the linear program converges to the ground truth (Figure~\ref{fig:WTcomputation}B, green line), while the runtime increases approximately as $\BigO(m^3)$ (purple line). However, for larger sample sizes the complexity will likely grow faster, as it is known that integer linear programs have exponential complexity~\cite{karp1972reducibility}. Note that the dimensionality of the feature space (in this example, it is two-dimensional) does not strongly influence runtime.

\subsection{Data}

\subsubsection{Simulated data}
\label{sec:HGM}

To demonstrate and validate our method, we simulated hierarchical datasets, i.e. we defined the \textit{hierarchical Gaussian mixture model}~(HGMM). Similar to a Gaussian mixture model, a HGMM has multiple classes from which units are drawn.  But here, each unit defines a Gaussian distribution with a unit-specific mean and covariance matrix. In each class, the unit means come from a class-specific Gaussian distribution, while the unit covariance matrices come from a class-specific Wishart distribution. 

\begin{defi}
	\label{HGM}
	Let $\N$ and $\W$ denote Gaussian and Wishart distributions respectively. A \textit{hierarchical Gaussian mixture model} is then defined by the number of classes~($K$), the number of units per class ($N_i$ for $i=1\ldots K$), the number of samples per unit ($M_j$ for $j=1\ldots\sum_{i=1}^K N_i$) and their feature dimensionality $F$, where
	\begin{itemize}
		\item each class $i$ is characterized by a Gaussian distribution $\N(\mu_i, \Sigma_i)$ with  $\mu_i\in~\mathbb{R}^{F}$ and $\Sigma_i\in \mathbb{R}^{F\times F}$ and a Wishart distribution $\W(n_i, \Lambda_i)$ with $n_i\geq F$ and $\Lambda_i \in \mathbb{R}^{F\times F}$;
		
		\item each unit $X_j$ belonging to a class $i$ is characterized by a Gaussian distribution $\N(\nu_j, \Gamma_j)$. Unit means are samples from the class-specific Gaussian $\nu_j \sim~\N(\mu_i,\Sigma_i)$ and unit covariance matrices are samples from the class-specific Wishart distribution $\Gamma_j \sim \W(n_i, \Lambda_i)$;
		
		\item the samples $S_k$ of each unit $j$ are iid distributed as $S_k \sim \N(\nu_j, \Gamma_j)$.
	\end{itemize}
\end{defi} 

\begin{figure}[t]
	\centering
	\includegraphics{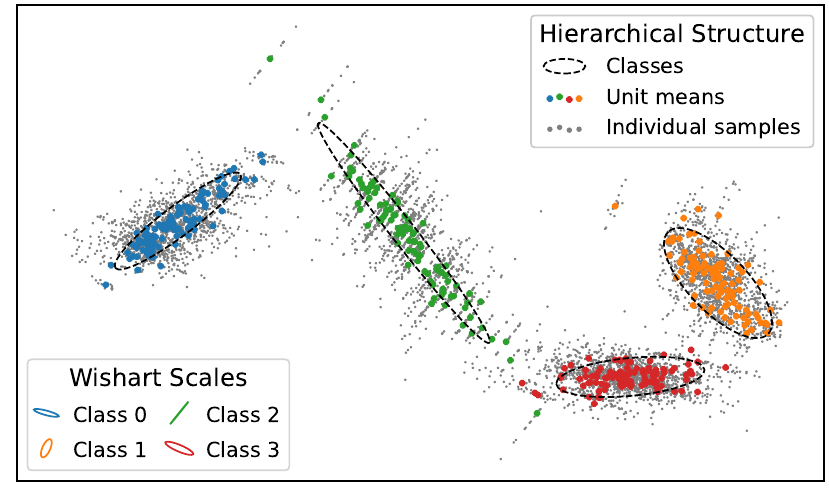}
	\caption{Hierarchical Gaussian mixture model (HGMM). This two-dimensional example dataset has $K=4$ classes with $N=100$ units each. The gray points show the samples from all units ($M=15$ samples per unit). Note that some of the units in the red and green classes have similar means, but their covariances are very different.}
	\label{fig:HGMclean}
\end{figure}

A HGMM is specified by the set of class-specific parameters $\{\mu_i, \Sigma_i, n_i, \Lambda_i\}$. For example, Figure~\ref{fig:HGMclean} shows a two-dimensional ($F=2$) HGMM with $K=4$ classes, $N=100$ units in each class, and $M=15$ samples in each unit. The class-specific Gaussian distributions (dashed black lines) are far away from each other because their means $\mu_i$ are chosen to be sufficiently different. The Within-class similarity of the unit means can be adjusted by $\Sigma_i$ (defining the shape of the dashed contours). Each class has its own Wishart scale matrix $\Lambda_i$; in this example, the peculiar Wishart scale of the green class makes green units easily distinguishable from the red units even when their means come close to the red class. Note, that the unit covariance matrices $\Gamma_j$ in any given class are not all the same. Their sampling process (from a Wishart distribution) is equivalent to drawing $n_i$ samples from a zero-centered Gaussian distribution with the Wishart scale as covariance matrix and estimating the sample covariance matrix. The larger the $n_i$, the closer all $\Gamma_j$ are to the respective $\Lambda_i$. We used $n_i=4$ in Figure~\ref{fig:HGMclean}. %In conclusion, the higher the degree of freedom $n_i$ of the class-specific Wishart distributions is, the more similar are the within-unit covariance matrices of the respective class.
 
\subsubsection{German election data}
\label{sec:GER}

The German parliamentary election was held in September 2017 with six major parties making it to the parliament. Germany is divided into 299 voting districts (\textit{Wahlkreise}). In each voting district, multiple polling stations (\textit{Wahlbezirke}) are set up. In our analysis we consider each voting district to be a unit with its polling stations being its samples.

The election data were directly downloaded from the \textit{Bundeswahlleiter} website (\url{https://tinyurl.com/mpevp355}). We removed results of all minor parties and normalized each polling station so that the percentages of the six major parties~---~CDU (including the Bavarian-only CSU), SPD, AfD, FDP, Grüne and Linke~---~sum to 1 (the feature dimension of each sample is therefore six). Voting by mail was counted in separate mail-only polling stations of the respective voting district. 

For this dataset, we a priori defined four classes: \textit{Cities} (all voting districts with population density of at least 1000 people per square kilometer; population densities (\url{https://tinyurl.com/3262nf8b}) also obtained from the \textit{Bundeswahlleiter} website), \textit{Southern Germany} (all districts in Bavaria and Baden-W\"urtemberg, excluding previously defined cities), \textit{Eastern Germany} (former DDR, excluding cities)  and \textit{Western Germany} (the rest).

\section{Results}

\subsection{Wasserstein $t$-SNE on simulated data}
\label{res:HGM}

To perform \textit{Wasserstein $t$-SNE}, we first compute the pairwise distance matrix between units in a dataset where each unit is considered to be a probability distribution over its samples. We then embed these units in 2D using the $t$-SNE algorithm.

\begin{figure}[t]
	\includegraphics{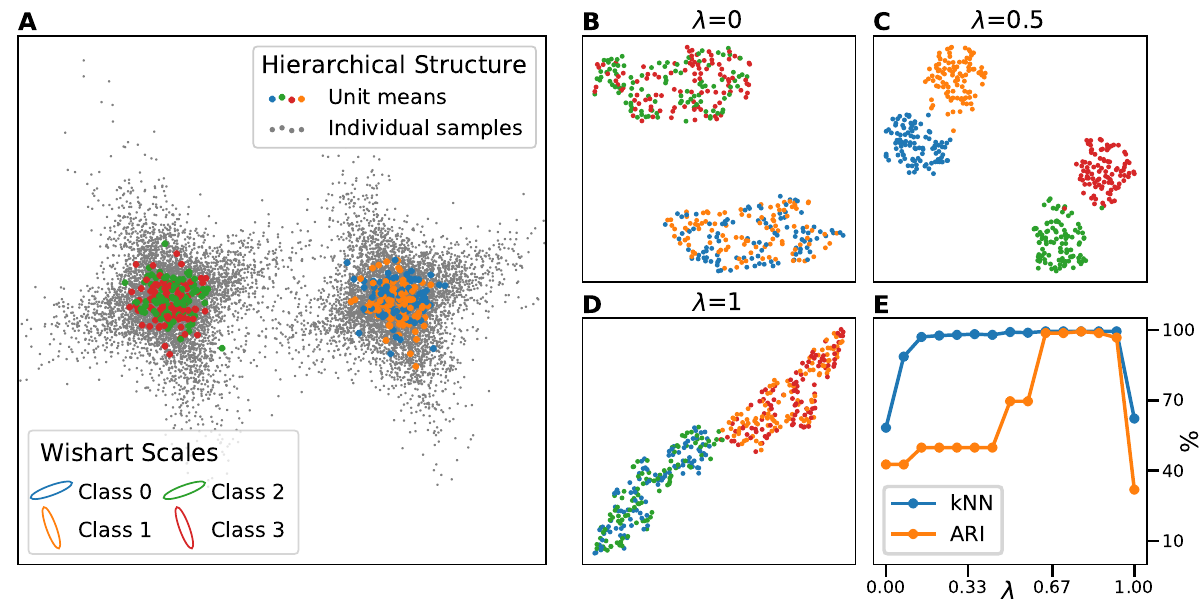}
	\caption{Wasserstein $t$-SNE. \abc{A}~This two-dimensional ($F=2$) HGMM was generated using $K=4$ classes with $N=100$ units each ($M=30$ samples per unit). Two pairs of classes have the same distribution of unit means, while two other pairs of classes have the same distribution of unit covariance matrices. \abc{B}~The mean-based  embedding ($\lambda=0$) is not able to separate some of the classes. \abc{C}~The Wasserstein embedding ($\lambda=0.5$) separates all four classes. \abc{D}~The covariance-based embedding ($\lambda=1$) is not able to separate some of the classes. \abc{E}~The performance at different values of $\lambda$ was assessed using the kNN accuracy ($k=5$) in the 2D embedding and the adjusted Rand index (ARI) obtained from Leiden clustering of the original distance matrix (kNN graph with $k=5$, resolution parameter $\gamma=0.08$).}
	\label{fig:HGMconcept}
\end{figure} 

Figure~\ref{fig:HGMconcept}A shows a two-dimensional ($F=2$) toy dataset that consists of $K=4$ classes, with $N=100$ units per class, and $M=30$ samples per unit. The red and  green classes have the same distribution of unit means; the same is true for the blue and the orange classes. Likewise, the red and orange classes have the same distribution of unit covariances; the same is true for the blue and the green classes. For a more detailed description see Section~\ref{sec:HGM}.

Depending on the value of $\lambda$ in (\ref{eq:lambda}), the resulting $t$-SNE embeddings show different structure. The mean-based embedding ($\lambda=0$) only separates the dataset into two clusters (Figure~\ref{fig:HGMconcept}B). Since it only takes the means into account, the orange class cannot be separated from the blue one, and the red class cannot be separated from the green one. Similarly, the covariance-based embedding ($\lambda=1$) only finds two clusters as well, mixing up blue with green and orange with red (Figure~\ref{fig:HGMconcept}D). In contrast, the Wasserstein embedding ($\lambda=0.5$) successfully separates all four classes from each other (Figure~\ref{fig:HGMconcept}C).

To measure the performance of the different $\lambda$ values, we used two different metrics. One metric is the k-nearest-neighbor (kNN) classification accuracy that measures the probability that a unit is labeled correctly by the majority vote of its $k=5$ nearest neighbors in the embedding (we used the \textsf{sklearn} implementation~\cite{goldberger2004neighbourhood}). The second metric is the adjusted Rand index~(ARI)~\cite{rand1971objective} that evaluates the agreement between the ground truth classes and the results of unsupervised clustering. We used the Leiden clustering algorithm~\cite{traag2019louvain}, applied to the Wasserstein distance matrix (here and below we used the \textsf{leidenalg} implementation~\cite{traag2019louvain} with resolution parameter $\gamma=0.08$ on the kNN graph built with $k=5$). Note that unlike the kNN accuracy, the ARI metric is independent of $t$-SNE.

Both metrics, kNN accuracy and ARI, peaked at $\lambda\in[0.7, 0.8]$ and showed markedly worse performance at both $\lambda=0$ and $\lambda=1$. Moreover, while the kNN accuracy was close to the peak at already $\lambda=0.5$, the ARI achieved higher values only for $\lambda>0.7$. The result indicates that putting more emphasis on the covariance structure helps the algorithm to cluster the data correctly. This shows the power of our generalized Wasserstein distance for Gaussian distributions, as in this case it outperforms the exact Wasserstein distance (corresponding to $\lambda=0.5$). 

%Another synthetic dataset which illustrates the domination of the means is shown in Figure~\ref{fig:HGMrandom}. Even though most classes are very close together, the kNN accuracies increase until $\lambda\approx0.93$ which might be an indication that the Wasserstein distance is dominated by the means. This effect will as well be a problem in experiments with real-world data that we do in the next section. Adjusting $\lambda$ corresponds to shifting the means closer together, which yields that their distances become less important.

%%% East vs. West examples
\begin{figure}[t]
	\includegraphics{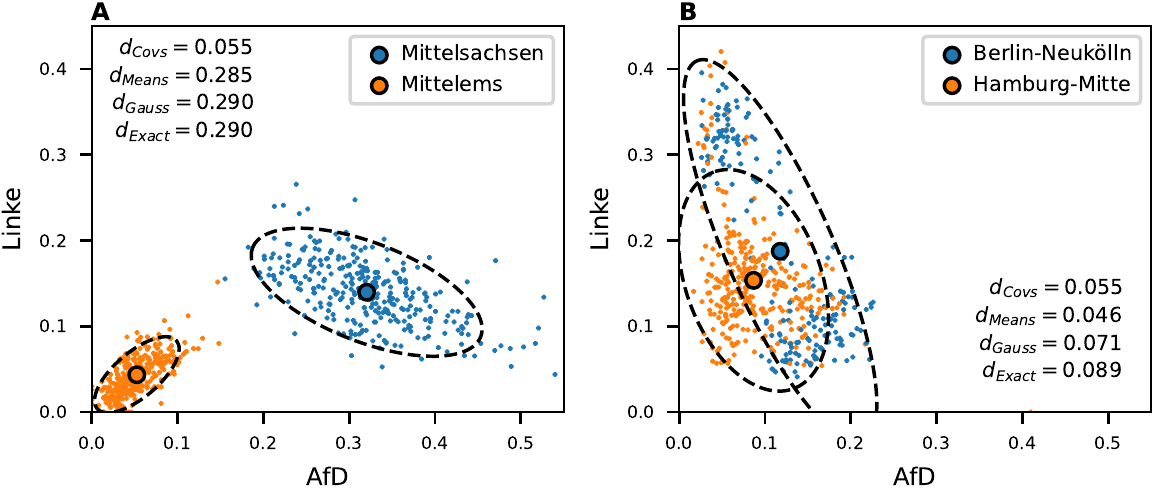}
	\caption{Example voting districts in the 2017 German parliamentary election. Four voting districts (units) are shown together with their respective polling stations (samples). \abc{A}~{Mittelems}, which is located in the rural western Germany, exhibits positive correlation between the votes for AfD and for Linke, whereas {Mittelsachsen} in eastern Germany shows negative correlation. 
	\abc{B}~The politically diverse district of Berlin-Neuk\"olln has bimodal structure in the within-unit distribution of AfD and Linke votes, whereas Hamburg-Mitte does not show such bimodality.}
	\label{fig:GERexample}
\end{figure}

\subsection{German parliamentary election 2017}
\label{res:GER}

\subsubsection{Gaussian Wasserstein $t$-SNE}

The dataset of the 2017 German parliamentary election dataset contains 299 voting districts (units), each having about 150--850 polling stations (samples). The samples are represented as a points in six-dimensional space (corresponding to six political parties), as described in Section~\ref{sec:GER}. For most units, the data could be reasonably well described by a multivariate Gaussian distribution (Figure~\ref{fig:GERexample}). For example, the Gaussian Wasserstein distance and the exact Wasserstein distance between Mittelsachsen and Mittelems districts, both equaled $0.290$ (Figure~\ref{fig:GERexample}A). For some districts the approximation was less good: e.g. between Berlin-Neuk\"olln and Hamburg-Mitte (Figure~\ref{fig:GERexample}B), the Gaussian Wasserstein distance was $0.071$ while the exact Wasserstein distance was $0.089$. This can be explained by the polarization of Berlin-Neuk\"olln, which had a bimodal structure that could not be captured by a multivariate normal distribution.  However, we found that most districts were well approximated by a Gaussian. 

%%% LEIDEN ALGORITHM GER
\begin{figure}[t]
	\includegraphics{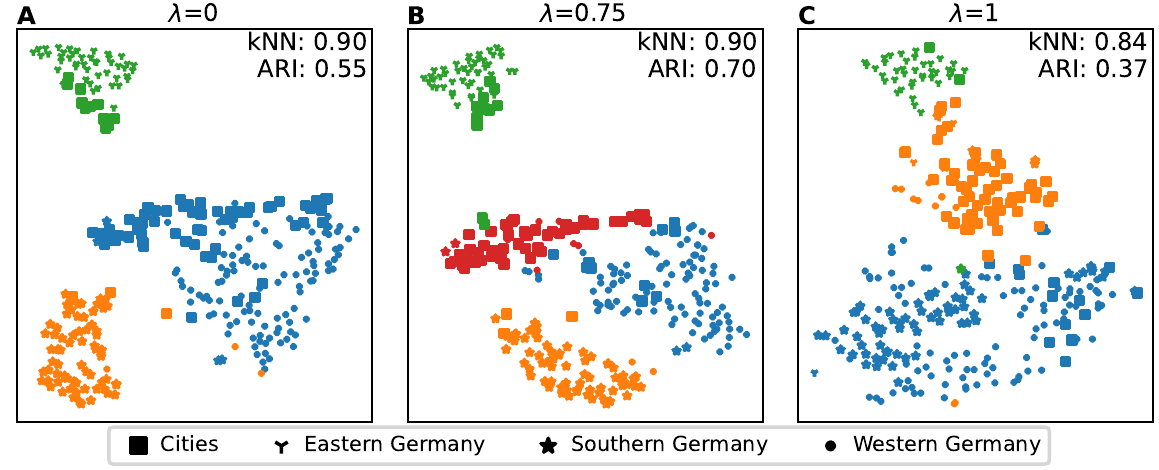}
	\caption{Wasserstein $t$-SNE of the 2017 German parliamentary election. For clustering, we used the Leiden algorithm (on the original distance matrix) with $k=5$ kNN graph and resolution parameter $\gamma=0.08$. Colors correspond to the Leiden clusters; marker shape corresponds to the a priori classes. \abc{A}~The mean-based embedding with $\lambda=0$ shows three clusters. \abc{B}~The Wasserstein embedding with $\lambda=0.75$ shows four clusters. \abc{C}~The covariance-based embedding with $\lambda=1$ shows three clusters.}
	\label{fig:GERembedding}
\end{figure}

We computed the pairwise Wasserstein distances between all pairs of units for different values of $\lambda$, and embedded the resulting distance matrices with the $t$-SNE algorithm (Figure~\ref{fig:GERembedding}). In parallel, we clustered each distance matrix by the Leiden algorithm (with resolution parameter $\gamma=0.08$ and kNN graph with $k=5$) and used the cluster assignments to color the $t$-SNE embeddings.  The Wasserstein embedding with $\lambda=0.75$ (Figure~\ref{fig:GERembedding}B) outperformed the mean-based ($\lambda=0$ and the covariance-based ($\lambda=1$) embeddings, and achieved a kNN accuracy of~$0.90$ and an ARI of~$0.70$. While the difference in the kNN accuracy was not very different for other values of $\lambda$ ($0.90$ for the mean-based and $0.82$ for the covariance-based embeddings), the difference in the ARI  was very pronounced (0.55 for the mean-based and 0.37 for the covariance-based embeddings). 

The ARI is sensitive to the number of clusters, which is not pre-specified in the Leiden algorithm. While it automatically found three clusters with $\lambda=0$ (Figure~\ref{fig:GERembedding}A) and $\lambda=1$ (Figure~\ref{fig:GERembedding}C), it found four clusters with $\lambda=0.75$, and these clusters corresponded well to the four classes (Cities, Western Germany, Eastern Germany, Southern Germany) that we defined a priori. 
In contrast, the mean-based distance matrix merged Western Germany with the Cities, and the covariance-based embedding  merged Western Germany with Southern Germany. This decreased the ARI for these embeddings, which is also visible in Figure~\ref{fig:GERexact}C. While the exact ARI values depend on the choice of $k$ and $\gamma$ parameters, our results show that the Wasserstein distances with $0<\lambda<1$ can be an improvement compared to ignoring either the information about the means or about the covariance structure.

%%% NORMAL GER Embedding
% \begin{figure}[t]
% 	\centering
% 	\includegraphics{GERembedding.pdf}
% 	\caption{Gaussian Wasserstein embeddings for the German Federal election. \abc{A}~The Euclidean embedding with $\lambda=0$ yields three clusters that separate the classes fairly well. \abc{B}~The Wasserstein embedding with $\lambda=0.75$ separates the \textit{Cities} from \textit{Western Germany}. \abc{C}~The Covariance embedding with $\lambda=1$ shows different clusters, where \textit{Western} and \textit{Southern Germany} become one cluster.}
% 	\label{fig:GERembedding}
% \end{figure}

\subsubsection{Covariance and correlation structure}

\begin{figure}[t]
	\includegraphics{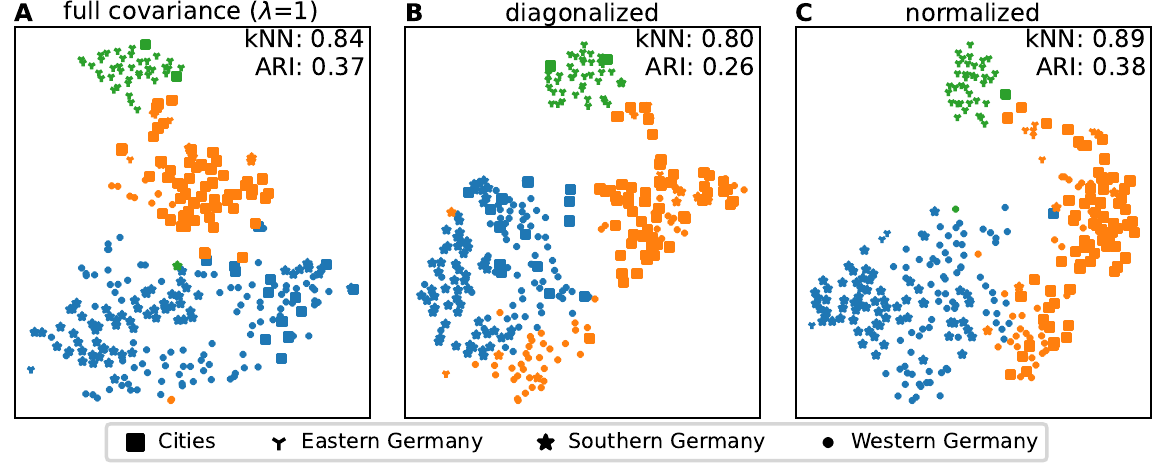}
	\caption{Variants of the covariance-based ($\lambda=1$) embedding. \abc{A}~The full covariance matrices were used to compute the pairwise distance matrix. The same embedding as in Figure~\ref{fig:GERembedding}C. \abc{B}~Only the diagonal entries of the covariance matrices were used, i.e. the marginal variances. \abc{C}~Each covariance matrix was normalized to become a correlation matrix.}
	\label{fig:GERcovariances}
\end{figure}

The covariance-based embedding clearly showed three clusters (Figure~\ref{fig:GERembedding}C), even though the means were completely ignored in this analysis. Where did this structure emerge from? Covariance is influenced by correlation and by marginal variances. To disentangle these two aspects of the data, we constructed a covariance-based embedding after zeroing out all off-diagonal values of all covariance matrices before computing the pairwise distance matrix (Figure~\ref{fig:GERcovariances}B). We also constructed a covariance-based embedding after normalizing all covariance matrices to be correlation matrices (Figure~\ref{fig:GERcovariances}C). Both embeddings were similar to the original covariance-based embedding (Figure~\ref{fig:GERcovariances}A), suggesting that there was meaningful information in marginal variances as well as in pairwise correlations.
%
%
%%% COVARIANCE ANALYSIS

%\subsubsection{Correlations}

\begin{figure}[t]
	\centering
	\includegraphics{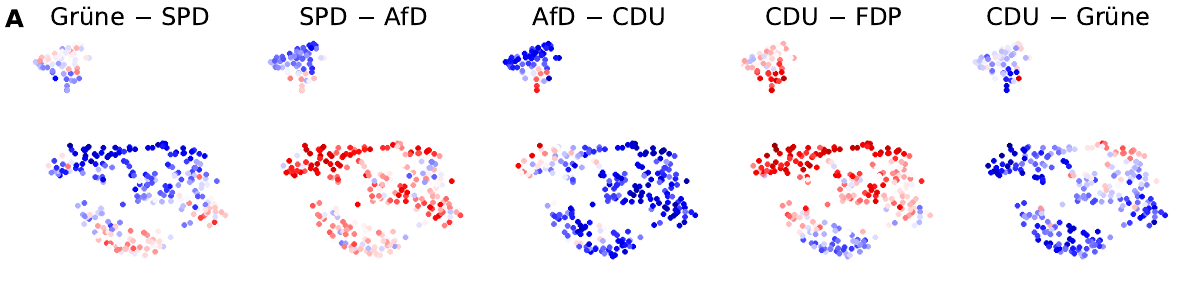}\\
	\includegraphics{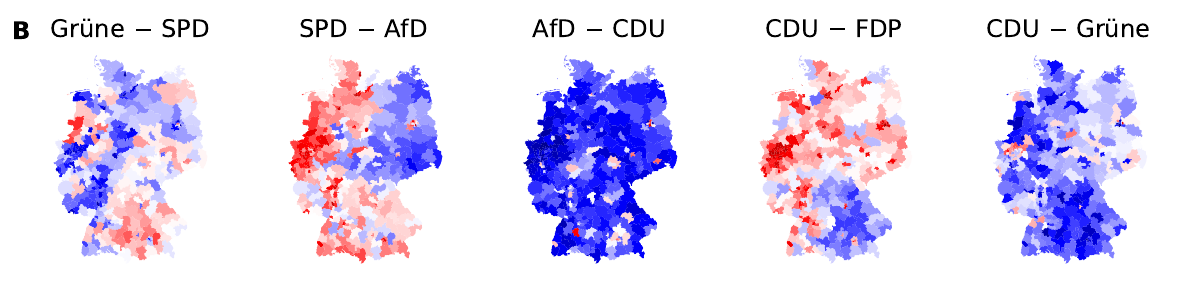}
	\caption{Pairwise correlations between parties. Colors represent the Pearson correlation coefficient of the respective parties in each voting district, from blue ($-1$) to red ($1$). We chose five (out of 15) pairs of parties showing the most interesting patterns. \abc{A} Party correlations overlayed over the Gaussian Wasserstein embedding with $\lambda=0.75$ (as in Figure~\ref{fig:GERembedding}B). \abc{B} Party correlations overlayed over the geographical map of Germany.}
	\label{fig:GERcorrelation}
\end{figure}

Figure~\ref{fig:GERcovariances}C demonstrates that class information was present in the within-unit correlations, and indeed we saw earlier that parties' results can correlate differently in different voting districts (Figure~\ref{fig:GERexample}). To visualize this effect, we overlayed pairwise correlation coefficients on the Gaussian Wasserstein embedding with $\lambda=0.75$ (Figure~\ref{fig:GERcorrelation}A). For several pairs of parties, correlation  strongly depended on the class, e.g. AfD and SPD were positively correlated in the Cities, but negatively correlated in Eastern Germany. This indicates that people who vote SPD in the cities tend to live in the same neighborhoods (i.e. same polling stations within a given district) as people who vote AfD, whereas in the rural east they tend to live in different neighborhoods. This suggests that these parties are perceived differently in different parts of the country, opposing each other in some of the regions but sharing sympathizers in others. Previous research has shown that many voters switched from {SPD} to {AfD} in the 2017 election~\cite{gortz2020sozialstrukturelle}. Our analysis indicates that this effect may have happened mostly in the east.

\subsubsection{Exact Wasserstein Embedding}

\begin{figure}[t]
	\centering
	\includegraphics{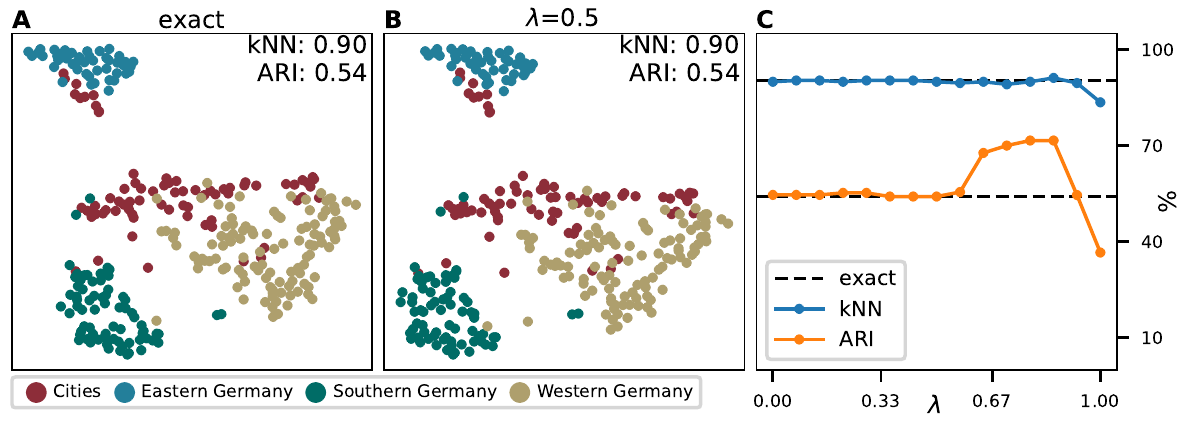}
	\caption{Comparison of the Wasserstein $t$-SNE embeddings based on the Gaussian approximation and based on the exact Wasserstein distances. \abc{A}~The exact Wasserstein $t$-SNE embedding separates the classes. \abc{B}~The Gaussian Wasserstein embedding with $\lambda=0.5$ shows similar structure. \abc{C}~ The kNN classification accuracy ($k=5$) and the ARI based on the Leiden clustering ($k=5$ kNN graph, resolution parameter $\gamma=0.08$) are shown for different values of $\lambda$. The black dashed lines show the kNN accuracy and the ARI of the exact Wasserstein embedding.}
	\label{fig:GERexact}
\end{figure}

To verify that the Gaussian approximation used above did not strongly distort the embedding, we also did an embedding based on the exact Wasserstein distance. As explained in Section~\ref{sec:linprog}, we calculated the exact Wasserstein distances using linear programming. While this approach is more faithful to the data, it requires much longer computation time; calculating  all $299 \cdot 298 / 2 = 44,551$ pairwise exact Wasserstein distances between units took 43~hours on a machine with 8~CPU cores at 3.0~GHz. The resulting embedding (Figure~\ref{fig:GERexact}A) was very similar to the Gaussian embedding with $\lambda=0.5$ (Figure~\ref{fig:GERexact}B). 
%One example of a minor difference was given by {Berlin-Neuk\"olln} district that was embedded next to the Cities by the Wasserstein approach, but to \textit{Eastern Germany} in the Euclidean embedding. An explanation can be found by the nearest neighbors of the district: while in the Euclidean distance matrix {Berlin-Neukölln} has as top-3 nearest neighbors {Berlin-Pankow}, {Potsdam} and {Berlin-Mitte} (all \textit{Eastern Germany}), the Wasserstein distance yields nearest neighbors  from the cluster of \textit{Cities}. These are small nuances in the embeddings that globally show the same structure.  

The Gaussian approximation has one important benefit beyond the faster runtime: namely, it allows to adjust the $\lambda$ parameter. As a function of $\lambda$, the kNN accuracy was nearly flat (Figure~\ref{fig:GERexact}C) but the ARI of the Leiden clustering peaked for $\lambda>0.6$, corresponding to the regime where the Leiden algorithm identifies four clusters in the data instead of three. At the same time, the performance of the exact Wasserstein embedding is similar to $\lambda=0.5$ and hence is worse (Figure~\ref{fig:GERexact}C, black dashed lines). This shows that, compared to the exact Wasserstein distances, the Gaussian approximation with the flexible $\lambda$ parameter can improve not only the runtime but also the final embedding.

%%%%%%%%%%%%%%%%%%%
%In research track, we welcome research articles in all fields of machine learning, knowledge discovery, and data mining. Following in the footsteps of ECML PKDD, we are looking for high-quality papers in terms of novelty, technical excellence, potential impact, and presentation clarity. Papers should demonstrate that they make a substantial contribution to the field (e.g, improve the state-of-the-art or provide a new theoretical insight).
%%%%%%%%%%%%%%%%%%%

\section{Discussion}
\label{discussion}

%%% HIGH-LEVEL SUMMARY
In this work we introduced \textit{Wasserstein $t$-SNE} as a method to visualize hierarchical datasets. The main idea is to compute the pairwise distance matrix between the units of interest, using the Wasserstein metric to compare the distribution of their samples; then we use $t$-SNE to make a 2D embedding of the resulting distance matrix (Figure~\ref{fig:WTcomic}). Using a simulated and a real-life dataset, we showed that our approach can outperform standard $t$-SNE based on unit averages (Figures~\ref{fig:HGMconcept}, \ref{fig:GERembedding}).

%%% MORE TECHNICAL SUMMARY
There are two different ways to use Wasserstein $t$-SNE. One way is to approximate each unit by a multivariate Gaussian distribution which is fast but not scalable to high feature dimensionality. The second way is to compute the exact Wasserstein distances which is more accurate but substantially slower if there are many samples. Both ways require to compute and store the pairwise distance matrix which gets unfeasible for very large datasets. A benefit of using the Gaussian approximation is that it allows to put emphasis on either the means or the covariances of the units by adjusting the $\lambda$ parameter in the distance definition~(\ref{eq:lambda}). We suggested this definition as a novel generalization of the closed-form solution for the 2-Wasserstein distance between Gaussian distributions. To compute the exact Wasserstein distances, we solve a linear program for each distance calculation. We developed an approach that scales with the number of samples and allows to compute Wasserstein distances between samples from continuous distributions. Using this approach, it took us $\sim$0.1~s to compute the Wasserstein distance between two Gaussian samples of size $n=100$ on a standard desktop computer, and $\sim$100~s for $n=1000$ (Figure~\ref{fig:WTcomputation}). The feature dimensionality of the Gaussians does not play a role here.

% To evaluate our method we initially looked at synthetic data, i.e., we designed datasets such that the covariances of their units contained information about the class membership. Using the Gaussian approximation we were able to show that our method clusters the classes correctly while straight-forward approaches, such as collapsing the units to their means, fail. This was quantified by computing the Leiden clusters on the pairwise distance matrix. The $t$-SNE embedding with $\lambda{=}0.5$ clearly separated the classes and the ($t$-SNE independent) adjusted Rand-Index on the true labels even increased until $\lambda{\approx}0.85$.

%%% ELECTION DATA APPLICATION
Electoral data provide ample opportunities for statistical analysis, such as e.g. statistical fraud detection~\cite{kobak2016integer,kobak2016statistical}. Here we used the publicly available data from the 2017 German parliamentary election to demonstrate that Wasserstein $t$-SNE can be useful for analysis of real-life datasets. Our method produced a 2D visualization (`map') of the 299 German voting districts (Figure~\ref{fig:GERembedding}B). This visualization exhibited four clusters, that were in good agreement with the known sociopolitical division of Germany that we defined a priori. Moreover, the correlation coefficient between political parties varied smoothly over the embedding (Figure~\ref{fig:GERcorrelation}), and in some cases even changed the sign. This showed that the information about within-unit distributions can be valuable to provide concise visualizations of political landscape in an unsupervised way. 

%%% RELATED WORK
%What other methods exist to embed hierarchical data? 
We are not aware of other methods specifically designed to visualize hierarchical data. The naive approach is to collapse units to their means. However, we showed that this can be suboptimal whenever there is meaningful information in the unit covariances (e.g.~Figure~\ref{fig:GERembedding}). An alternative is to append some of the covariance-based features to the unit means. This approach was, e.g., used in the Wisconsin Breast Cancer dataset~\cite{data1992wsdm} (popularized by its role as a UCI benchmark) where samples are cells and units are the respective patients. Here the variance of each feature (such as cell radius or cell smoothness) was appended to the dataset as an additional separate feature. While this allows to use some of the covariance information, it removes all information about feature correlation. Finally, it is possible to base all the analysis on the sample level, instead of the unit level. Such a `sample-based' $t$-SNE embedding would show many more points than a `unit-based' $t$-SNE embedding. However, for datasets like the one shown in Figure~\ref{fig:HGMconcept} this would not result in a useful visualization, as it would yield only two clusters and not four (as there are two pairs of classes with strongly overlapping distribution of samples within the units). 

%%% CONCLUSION AND POSSIBLE APPLICATIONS
In summary, we believe that Wasserstein $t$-SNE is a promising method to visualize hierarchical datasets. Our results on synthetic data and on the 2017 German election data demonstrate that Wasserstein $t$-SNE can outperform standard alternatives and uncover meaningful structure in the data. We hope that our method can be useful in various domains. For example, social science often deals with hierarchical datasets, such as the European Values Study~\cite{data2017evs} where geographical regions can be seen as units while individual participants of the survey can be seen as samples. We believe that Wasserstein $t$-SNE can also be useful beyond the social and political science, e.g. for biomedical data.

\subsubsection{Acknowledgements}
We thank Philipp Berens for comments and suggestions. This research was funded by the Deutsche Forschungsgemeinschaft (Excellence Cluster 2064 ``Machine Learning: New Perspectives for Science'', 390727645), and by the German Ministry of Education and Research through the T\"ubingen AI Center (01IS18039A).

%%% Bibliography
 %\bibliographystyle{splncs04}
 %\bibliography{references}

\end{document}